\documentclass{article} 

\newif\ifprintsm
\printsmtrue

\usepackage[accepted,nohyperref]{icml2019}

\usepackage[utf8]{inputenc} 
\usepackage[T1]{fontenc}      
\usepackage{url, times}           
\usepackage{booktabs}           
\usepackage{amsfonts}           
\usepackage{nicefrac}       
\usepackage{microtype}      
\usepackage{subfigure}
\usepackage{xcolor,graphicx,bm,enumitem}
\usepackage{amsmath,amssymb,amsfonts,mathtools}
\usepackage{algorithm,algorithmic}

\newcommand{\inv}{^{\raisebox{.2ex}{$\scriptscriptstyle-1$}}}

\newcommand{\nnw}{{\tt w}}
\newcommand{\env}{{\mathcal{D}}}

\icmltitlerunning{Remember and Forget for Experience Replay}

\begin{document}
\twocolumn[
\icmltitle{Remember and Forget for Experience Replay}
	
\begin{icmlauthorlist}
\icmlauthor{Guido Novati}{ethz}
\icmlauthor{Petros Koumoutsakos}{ethz}
\end{icmlauthorlist}
\icmlaffiliation{ethz}{Computational Science \& Engineering Laboratory, ETH Zurich, Zurich, Switzerland}
\icmlcorrespondingauthor{Guido Novati}{novatig@ethz.ch}
\icmlcorrespondingauthor{Petros Koumoutsakos}{petros@ethz.ch}
	
\vskip 0.3in
]
\printAffiliationsAndNotice{} 

\begin{abstract}

Experience replay (ER) is a fundamental  component of off-policy deep reinforcement learning (RL). ER recalls experiences from past iterations to compute gradient estimates for the current policy, increasing data-efficiency. However, the accuracy of such updates may deteriorate when the policy diverges from past behaviors and can undermine the performance of ER.
Many algorithms mitigate this issue by tuning hyper-parameters to slow down policy changes. 
An alternative is to actively enforce the similarity between policy and the  experiences in the replay memory.
We introduce {\it Remember and Forget Experience Replay} ({\tt ReF-ER}), a novel method that can enhance RL algorithms with parameterized policies.
{\tt ReF-ER}
(1) skips gradients computed from experiences that are too unlikely with the current policy 
and 
(2) regulates policy changes within a trust region of the replayed behaviors. 
We couple {\tt ReF-ER} with Q-learning, deterministic policy gradient and off-policy gradient methods.
We find that  {\tt ReF-ER} consistently improves the performance of continuous-action, off-policy RL on  fully observable benchmarks and partially observable flow control problems.
\end{abstract}

\section{Introduction}
Deep reinforcement learning (RL) has an ever increasing number of success stories ranging from realistic simulated environments~\citep{schulman2015a,mnih2016}, robotics~\citep{levine2016} and games~\citep{mnih2015, silver2016}.
Experience Replay (ER)~\citep{lin1992} enhances RL algorithms by using information collected in past policy ($\mu$) iterations to compute updates for the current policy ($\pi$). ER has become one of the mainstay techniques to improve the sample-efficiency of off-policy deep RL. Sampling from a replay memory (RM) stabilizes stochastic gradient descent (SGD) by disrupting temporal correlations and extracts information from useful experiences over multiple updates~\citep{schaul2015a}. However, when $\pi$ is parameterized by a neural network (NN), SGD updates may result in significant changes to the policy, thereby shifting the distribution of states observed from the environment.
In this case sampling the RM for further updates may lead to incorrect gradient estimates, therefore deep RL methods must account for and limit the dissimilarity between $\pi$ and behaviors in the RM.  Previous works employed trust region methods to bound policy updates~\citep{schulman2015a, wang2016}. Despite several successes, deep RL algorithms are known to suffer from instabilities and exhibit high-variance of outcomes~\citep{islam2017, henderson2017}, especially continuous-action methods employing the stochastic \citep{sutton2000} or deterministic~\citep{silver2014} policy gradients (PG or DPG).

In this work we redesign ER in order to control the similarity between the replay behaviors $\mu$ used to compute updates and the policy $\pi$. More specifically, we classify experiences either as ``near-policy" or ``far-policy", depending on the ratio $\rho$ of probabilities of selecting the associated action with $\pi$ and that with $\mu$.
The weight $\rho$ appears in many estimators that are used with ER such as the off-policy policy gradients (off-PG)~\citep{degris2012} and the off-policy return-based evaluation algorithm Retrace~\citep{munos2016}. Here we propose and analyze Remember and Forget Experience Replay ({\tt ReF-ER}), an ER method that can be applied to any off-policy RL algorithm with parameterized policies. {\tt ReF-ER} limits the fraction of far-policy samples in the RM, and computes gradient estimates only from near-policy  experiences. Furthermore, these hyper-parameters can be gradually annealed during training to obtain increasingly accurate updates from nearly on-policy experiences. We show that {\tt ReF-ER} allows better stability and performance than conventional ER in all three main classes of continuous-actions off-policy deep RL algorithms: methods based on the DPG (ie. {\tt DDPG}~\citep{lillicrap2015}), methods based on Q-learning (ie. {\tt NAF}~\citep{gu2016}), and with off-PG~\citep{degris2012,wang2016}.

In recent years, there is a growing interest in coupling RL with high-fidelity physics simulations~\citep{reddy2016, novati2017,colabrese2017, verma2018}.
The computational cost of these simulations calls for reliable and data-efficient RL methods that do not require problem-specific tweaks to the hyper-parameters (HP).
Moreover, while on-policy training of simple architectures has been shown to be sufficient in some benchmarks~\citep{rajeswaran2017}, agents aiming to solve complex problems with partially observable dynamics might require deep or recurrent models that can be trained more efficiently with off-policy methods. We analyze {\tt ReF-ER} on the OpenAI Gym~\citep{brockman2016} as well as fluid-dynamics simulations to show that it reliably obtains competitive results without requiring extensive HP optimization.

\section{Methods}\label{sec:pre}
Consider the sequential decision process of an agent aiming to optimize its interaction with the environment. 
At each step $t$, the agent observes its state $s_t \in \mathbb{R}^{d_S}$, performs an action by sampling a policy $a_t \sim \mu_t(a | s_t) \in \mathbb{R}^{d_A}$, and transitions to a new state $s_{t+1} \sim \env(s | a_t, s_t)$ with reward $r_{t+1}\in \mathbb{R}$. The experiences $\{s_t, a_t, r_t, \mu_t\}$ are stored in a RM, which constitutes the data used by off-policy RL to train the parametric policy $ \pi^\nnw(a | s)$.
The importance weight $\rho_t {=} \pi^\nnw(a_t | s_t)/ \mu_t(a_t | s_t)$ is the ratio between the probability of selecting $a_t$ with the current $\pi^\nnw$ and with the behavior $\mu_t$, which gradually becomes dissimilar from $\pi^\nnw$ as the latter is trained.
The on-policy state-action value $Q^{\pi^\nnw}(s, a)$ measures the expected returns from $(s, a)$ following $\pi^\nnw$:
\begin{equation}
Q^{\pi^\nnw}(s,a) = ~\mathrel{\raisebox{4pt}{$ \mathop{\mathbb{E}}_{\substack{s_t\sim\env \\ a_{t}\sim\pi^\nnw}} $}}
\left[\left. \sum_{{\scriptscriptstyle t=0}}^{{\scriptscriptstyle \infty}} \gamma^t r_{t+1} ~\right| s_0{=}s,~a_0{=}a\right]
\end{equation}
Here $\gamma$ is a discount factor. The value of state $s$ is the on-policy expectation $V^{\pi^\nnw}(s) = \mathbb{E}_{a\sim{\pi^\nnw}}\left[ Q^{\pi^\nnw}(s,a) \right]$. 
In this work we focus on three deep-RL algorithms, each representing one class of off-policy continuous action RL methods. 

\textbf{DDPG}~\citep{lillicrap2015} is a method based on deterministic PG which trains two networks by ER. The value-network (a.k.a. critic) outputs $Q^{{\tt w}'}(s,a)$ and is trained to minimize the L2 distance from the temporal difference (TD) target $\hat{Q}_t = r_{t+1} + \gamma \mathop{\mathbb{E}}_{a' \sim \pi} {Q^{\nnw'}(s_{t+1}, a') }$:
\begin{equation}
\mathcal{L}^\text{Q}(\nnw') =\frac{1}{2}
~\mathrel{\raisebox{3pt}{$ \mathop{\mathbb{E}}_{\substack{s_k\sim B(s) \\ a_k \sim \mu_k}} $}}
\left[ Q^{\tt w'}(s_k, a_k) -\hat{Q}_k \right]^2 \label{eq:qltgt}
\end{equation}
Here $B(s) \propto \sum_{k=t{-}N}^t P(s_k {=} s~|~s_0,~a_k {\sim} \mu_k)$ is the probability of sampling state $s$ from a RM containing the last $N$ experiences of the agent acting with policies $\mu_k$. The policy-network is trained to output actions $\mathbf{m}^\nnw$ that maximize the returns predicted by the critic~\citep{silver2014}:
\begin{equation}
\mathcal{L}^\text{DPG}(\nnw) = - 
\mathrel{\raisebox{1pt}{$ \mathop{\mathbb{E}}_{s_k\sim B(s)} $}}
 \left[ Q^{\tt w'}(s_k, \mathbf{m}^\nnw (s_k)) \right] \label{eq.dpg}
\end{equation}

\textbf{NAF}~\citep{gu2016} is the state-of-the-art of Q-learning based algorithms for continuous-action problems. It employs a quadratic-form approximation of the $Q^\nnw$ value:
\begin{equation} 
\cramped{Q^\nnw_\text{NAF}(s,a) =V^\nnw(s) - \left[ a {-} \mathbf{m}^\nnw(s) \right]^\intercal \mathbf{L}^\nnw \left[\mathbf{L}^\nnw\right]^\intercal \left[ a {-}  \mathbf{m}^\nnw(s) \right]}
\end{equation}
Given a state $s$, a single network estimates its value $V^\nnw$, the optimal action $\mathbf{m}^\nnw(s)$, and the lower-triangular matrix $\mathbf{L}^\nnw(s)$ that parameterizes the advantage. Like {\tt DDPG}, {\tt NAF} is trained by ER with the Q-learning target (Eq.~\ref{eq:qltgt}).
For both {\tt DDPG} and {\tt NAF}, we include exploratory Gaussian noise in the policy $\pi^\nnw  = \mathbf{m}^\nnw {+} \mathcal{N}(\mathbf{0}, \sigma^2 \mathbf{I})$ with $\sigma{=}0.2$ (to compute $\rho_t$ or the Kullback-Leibler divergence $D_{KL}$ between policies).

\textbf{V-RACER} is the method we propose to analyze off-policy PG (off-PG) and ER. Given $s$, a single NN outputs the value $V^\nnw$, the mean $\mathbf{m}^\nnw$ and diagonal covariance $\boldsymbol{\Sigma}^\nnw$ of the Gaussian policy $\pi^\nnw(a | s)$.
The policy is updated with the off-policy objective~\citep{degris2012}:
\begin{equation}
\mathcal{L}^\text{off-PG}(\nnw) ={-}
\mathrel{\raisebox{1pt}{$ \mathop{\mathbb{E}}_{\substack{s_k\sim B(s) \\ a_k \sim \mu_k}} $}} 
\left[ \rho_k \left(\hat{Q}^\text{ret}_k - V^\nnw(s_{k})\right)  \right]
\end{equation}
On-policy returns are estimated with Retrace~\citep{munos2016}, which takes into account rewards obtained by $\mu_t$:
\begin{equation}
\hat{Q}^\text{ret}_{t-1} = r_{t} +\gamma V^\nnw(s_{t})+\gamma  \bar{\rho}_{t} \left[ \hat{Q}^\text{ret}_{t} {-} Q^\nnw(s_{t},a_{t})\right] \label{eq:qret}
\end{equation}
Here we defined $\bar{\rho}_{t} {=} \min \{1, \rho_{t}\}$. {\tt V-RACER} avoids training a NN for the action value by approximating $Q^\nnw{:=}V^\nnw$ (i.e. it assumes that any individual action has a small effect on returns~\citep{tucker2018}). 
The on-policy state value is estimated with the ``variance truncation and bias correction trick'' (TBC)~\citep{wang2016}:
\begin{equation}
\hat{V}^\text{tbc}_t = V^\nnw(s_t) + \bar{\rho}_{t} [\hat{Q}^\text{ret}_{t} - Q^\nnw(s_{t},a_{t})] \label{eq:tbc}
\end{equation}
From Eq.~\ref{eq:qret} and~\ref{eq:tbc} we obtain $\hat{Q}^\text{ret}_t {=} r_{t{+}1} {+} \gamma \hat{V}^\text{tbc}_{t{+}1}$. From this, Eq.~\ref{eq:tbc} and $Q^\nnw{:=}V^\nnw$, we obtain a recursive estimator for the on-policy state value that depends on $V^\nnw$ alone:
\begin{equation}
\hat{V}^\text{tbc}_t = V^\nnw(s_t) +  \bar{\rho}_{t} \left[ r_{t+1} + \gamma \hat{V}^\text{tbc}_{t{+}1} -  V^\nnw(s_t) \right] \label{eq:vtbc}
\end{equation}
This target is equivalent to the recently proposed V-trace estimator~\citep{espeholt2018} when all importance weights are clipped at 1, which was empirically found by the authors to be the best-performing solution. Finally, the value estimate is trained to minimize the loss:
\begin{equation}
\mathcal{L}^\text{tbc}(\nnw) =\frac{1}{2}
\mathrel{\raisebox{3pt}{$ \mathop{\mathbb{E}}_{\substack{s_k\sim B(s) \\ a_k \sim \mu_k}} $}} 
\left[ V^\nnw(s_t) -\hat{V}^\text{tbc}_t \right]^2 \label{eq:rettgt}
\end{equation}


In order to estimate $\hat{V}^\text{tbc}_t$ for a sampled time step $t$, Eq.~\ref{eq:vtbc} requires $V^\nnw$ and $\rho_t$ for all following steps in sample $t$'s episode. These are naturally computed when training from batches of episodes (as in {\tt ACER}~\citep{wang2016}) rather than time steps (as in {\tt DDPG} and {\tt NAF}). However, the information contained in consecutive steps is correlated, worsening the quality of the gradient estimate, and episodes may be composed of thousands of time steps, increasing the computational cost. 
To efficiently train from uncorrelated time steps, {\tt V-RACER} stores for each sample the most recently computed estimates of $V^\nnw(s_k)$, $\rho_k$ and $\hat{V}^\text{tbc}_k$. When a time step is sampled, the stored $\hat{V}^\text{tbc}_k$ is used to compute the gradients. At the same time, the current NN outputs are used to update $V^\nnw(s_k)$, $\rho_k$ and to correct $\hat{V}^\text{tbc}$ for all prior time-steps in the episode with Eq.~\ref{eq:vtbc}. Each algorithm and the remaining implementation details are described in App.~A.

\section{Remember and Forget Experience Replay}\label{sec:ER}
In off-policy RL it is common to maximize on-policy returns estimated over the distribution of states contained in a RM. In fact, each method introduced in Sec.~\ref{sec:pre} relies on computing estimates over the distribution $B(s)$ of states observed by the agent following behaviors $\mu_k$ over prior steps $k$.
However, as $\pi^\nnw$ gradually shifts away from previous behaviors, $B(s)$ is increasingly dissimilar from the on-policy distribution, and trying to increase an off-policy performance metric may not improve on-policy outcomes. 
This issue can be compounded with algorithm-specific concerns. For example, the dissimilarity between $\mu_k$ and $\pi^\nnw$ may cause vanishing or diverging importance weights $\rho_k$, thereby increasing the variance of the off-PG and deteriorating the convergence speed of Retrace (and V-trace) by inducing ``trace-cutting''~\citep{munos2016}. Multiple remedies have been proposed to address these issues. For example, {\tt ACER} tunes the learning rate and uses a target-network~\citep{mnih2015}, updated as a delayed copy of the policy-network, to constrain policy updates. Target-networks are also employed in {\tt DDPG} to slow down the feedback loop between value-network and policy-network optimizations. This feedback loop causes overestimated action values that can only be corrected by acquiring new on-policy samples. Recent works~\citep{henderson2017} have shown the opaque variability of outcomes of continuous-action deep RL algorithms depending on hyper-parameters. Target-networks may be one of the sources of this unpredictability. 
In fact, when using deep approximators, there is no guarantee that the small weight changes imposed by target-networks correspond to small changes in the network's output.

This work explores the benefits of actively managing the ``off-policyness'' of the experiences used by ER. We propose a set of simple techniques, collectively referred to as Remember and Forget ER ({\tt ReF-ER}), that can be applied to any off-policy RL method with parameterized policies.

\begin{itemize}[leftmargin=*]
\item The cost functions are minimized by estimating the gradients $\hat{\mathbf{g}}$ with mini-batches of experiences drawn from a RM.
We compute the importance weight $\rho_t$ of each experience and classify it as ``near-policy" if $1/c_{\text{max}} {<}\rho_t {<} c_{\text{max}}$ with $c_{\text{max}}{>}1$. Samples with vanishing ($\rho_t{<}1/c_{\text{max}}$) or exploding ($\rho_t{>}c_{\text{max}}$) importance weights are classified as ``far-policy". When computing  off-policy estimators with finite batch-sizes, such as $\hat{Q}^\text{ret}$ or the off-PG, ``far-policy" samples may either be irrelevant or increase the variance. For this reason, \textbf{(Rule 1:) the gradients computed from far-policy samples are clipped to zero.} In order to efficiently approximate the number of far-policy samples in the RM,  we store for each step its most recent $\rho_t$.


\item \textbf{(Rule 2:) Policy updates are penalized in order to attract the current policy $\pi^\nnw$ towards past behaviors}:
	\begin{equation}\label{eq.penaliz}
	\hat{\mathbf{g}}^{\text{ReF-ER}}(\nnw) {=} \begin{cases}
		\beta \hat{\mathbf{g}}(\nnw) ~{-} (1{-}\beta) \hat{\mathbf{g}}^{D}(\nnw) & \text{if } \frac{1}{c_{\text{max}}} {<}\rho_t {<} c_{\text{max}} \\
		\hspace{0.98cm}{-} (1{-}\beta) \hat{\mathbf{g}}^{D}(\nnw)  & \text{otherwise}
	\end{cases}
	\end{equation}
Here we penalize the ``off-policyness'' of the RM with:
	\begin{equation}
		\hat{\mathbf{g}}^{D}(\nnw) =\mathbb{E}_{s_k\sim B(s) } \left[\nabla  D_{\text{KL}} \left( \mu_k \| \pi^\nnw(\cdot | s_k) \right)  \right]
	\end{equation}
	The coefficient $\beta \in [0, 1]$ is updated at each step such that a set fraction $D \in (0, 1)$ of samples are far-policy:
	\begin{equation}
	\beta \leftarrow \begin{cases}
	(1-\eta) \beta & \text{if } n_\text{far} / N  > D\\
	(1-\eta)\beta + \eta,              & \text{otherwise}
	\end{cases}\label{eq:penl}
	\end{equation}
	Here $\eta$ is the NN's learning rate, $N$ is the number of experiences in the RM, of which $n_\text{far}$ are far-policy. Note that iteratively updating $\beta$ with Eq.~\ref{eq:penl} has fixed points in $\beta{=}0$ for $ n_\text{far} /N {>}D$ and in $\beta{=}1$ otherwise. 
\end{itemize}
We remark that alternative metrics of the relevance of training samples were considered, such as $D_{KL}$ or its discounted cumulative sum, before settling on the present formulation. {\tt ReF-ER} aims to reduce the sensitivity on the NN architecture and HP by controlling the rate at which the policy can deviate from the replayed behaviors. For $c_{\text{max}}{\rightarrow} 1$ and $D{\rightarrow} 0$, {\tt ReF-ER} becomes asymptotically equivalent to computing updates from an on-policy dataset.
Therefore, we anneal {\tt ReF-ER}'s $c_\text{max}$ and the NN's learning rate according to:
\begin{equation}
c_\text{max}(t) = 1 {+} C/(1 + A\cdot t), ~~ \eta(t) = \eta / (1 + A \cdot t) \label{eq:annl}
\end{equation}
Here $t$ is the time step index, $A$ regulates annealing, and $\eta$ is the initial learning rate. $c_\text{max}$ determines how much $\pi^\nnw$ is allowed to differ from the replayed behaviors. By annealing $c_\text{max}$ we allow fast improvements at the beginning of training, when inaccurate policy gradients might be sufficient to estimate a good direction for the update. Conversely, during the later stages of training, precise updates can be computed from almost on-policy samples.  We use $A{=}5{\cdot}10^{{-}7}$, $C{=}4$, $D{=}0.1$, and $N{=}2^{18}$ for all results with {\tt ReF-ER} in the main text. The effect of these hyper parameters is further discussed in a detailed sensitivity analysis reported in the Supplementary Material.

\section{Related work}\label{sec:related}
The rules that determine which samples are kept in the RM and how they are used for training can be designed to address specific objectives. For example, it may be necessary to properly plan ER to prevent lifelong learning agents from forgetting previously mastered tasks~\citep{isele2018}. ER can be used to train transition models in planning-based RL~\citep{pan2018}, or to help shape NN features by training off-policy learners on auxiliary tasks~\citep{schaul2015b,jaderberg2016b}. When rewards are sparse, RL agents can be trained to repeat previous outcomes~\citep{andrychowicz2017} or to reproduce successful states or episodes~\citep{oh2018,goyal2018}.

In the next section we compare {\tt ReF-ER} to conventional ER and \textit{Prioritized Experience Replay}~\citep{schaul2015a} ({\tt PER}). {\tt PER} improves the performance of {\tt DQN}~\citep{mnih2015} by biasing sampling in favor of experiences that cause large temporal-difference (TD) errors. TD errors may signal rare events that would convey useful information to the learner. 
\citet{de2015} proposes a modification to ER that increases the diversity of behaviors contained in the RM, which is the opposite of what {\tt ReF-ER} achieves. Because the ideas proposed by \citet{de2015} cannot readily be applied to complex tasks (the authors state that their method is not suitable when the policy is advanced for many iterations), we compare {\tt ReF-ER} only to {\tt PER} and conventional ER. We assume that if increasing the diversity of experiences in the RM were beneficial to off-policy RL then either {\tt PER} or ER would outperform {\tt ReF-ER}.

{\tt ReF-ER} is inspired by the techniques developed for on-policy RL to bound policy changes in {\tt PPO}~\citep{schulman2017}. Rule 1 of {\tt ReF-ER} is similar to the clipped objective function of {\tt PPO} (gradients are zero if $\rho_t$ is outside of some range). However, Rule 1 is not affected by the sign of the advantage estimate and clips both policy and value gradients. Another variant of {\tt PPO} penalizes $D_{KL}(\mu_t || \pi^\nnw)$ in a similar manner to Rule 2 (also~\citet{schulman2015a} and~\citet{wang2016} employ trust-region schemes in the on- and off-policy setting respectively). {\tt PPO} picks one of the two techniques, and the authors find that gradient-clipping performs better than penalization. Conversely, in {\tt ReF-ER} Rules 1 and 2 complement each other and can be applied to most off-policy RL methods with parametric policies. 

{\tt V-RACER} shares many similarities with {\tt ACER}~\citep{wang2016} and {\tt IMPALA}~\citep{espeholt2018} and is a secondary contribution of this work. The improvements introduced by {\tt V-RACER} have the purpose of aiding our analysis of {\tt ReF-ER}: (1) {\tt V-RACER} employs a single NN; not requiring expensive architectures eases reproducibility and exploration of the HP (e.g. continuous-{\tt ACER} uses 9 NN evaluations per gradient). (2) {\tt V-RACER} samples time steps rather than episodes (like {\tt DDPG} and {\tt NAF} and unlike {\tt ACER} and {\tt IMPALA}), further reducing its cost (episodes may consist of thousands of steps). (3)  {\tt V-RACER} does not introduce techniques that would interfere with {\tt ReF-ER} and affect its analysis. Specifically, {\tt ACER} uses the TBC (Sec.~\ref{sec:pre}) to clip policy gradients, employs a target-network to bound policy updates with a trust-region scheme, and modifies Retrace to use $\sqrt[{d_A}]{\rho}$ instead of $\rho$. Lacking these techniques, we expect {\tt V-RACER} to require {\tt ReF-ER} to deal with unbounded importance weights. Because of points (1) and (2), the computational complexity of {\tt V-RACER} is approximately two orders of magnitude lower than that of {\tt ACER}.

\section{Results}
In this section we couple {\tt ReF-ER}, conventional ER and {\tt PER} with one method from each of the three main classes of deep continuous-action RL algorithms: {\tt DDPG}, {\tt NAF}, and {\tt V-RACER}. In order to separate the effects of its two components, we distinguish between {\tt ReF-ER-1}, which uses only Rule 1, {\tt ReF-ER-2}, using only Rule 2, and the full {\tt ReF-ER}. The performance of each combination of algorithms is measured on the MuJoCo~\citep{todorov2012} tasks of OpenAI Gym~\citep{brockman2016} by plotting the mean cumulative reward $R= \sum_t r_{t}$. Each plot tracks the average $R$ among all episodes entering the RM within intervals of $2{\cdot}10^5$ time steps averaged over five differently seeded training trials. For clarity, we highlight the contours of the $20^{th}$ to $80^{th}$ percentiles of $R$ only of the best performing alternatives to the proposed methods. The code to reproduce all present results is available on GitHub.\footnote{ \url{https://github.com/cselab/smarties}}

\subsection{Results for {\tt DDPG}}\label{sec:ddpgret}
\begin{figure*}[t] \centering
	\includegraphics[width=\textwidth, trim={0 0 0 0}, clip]{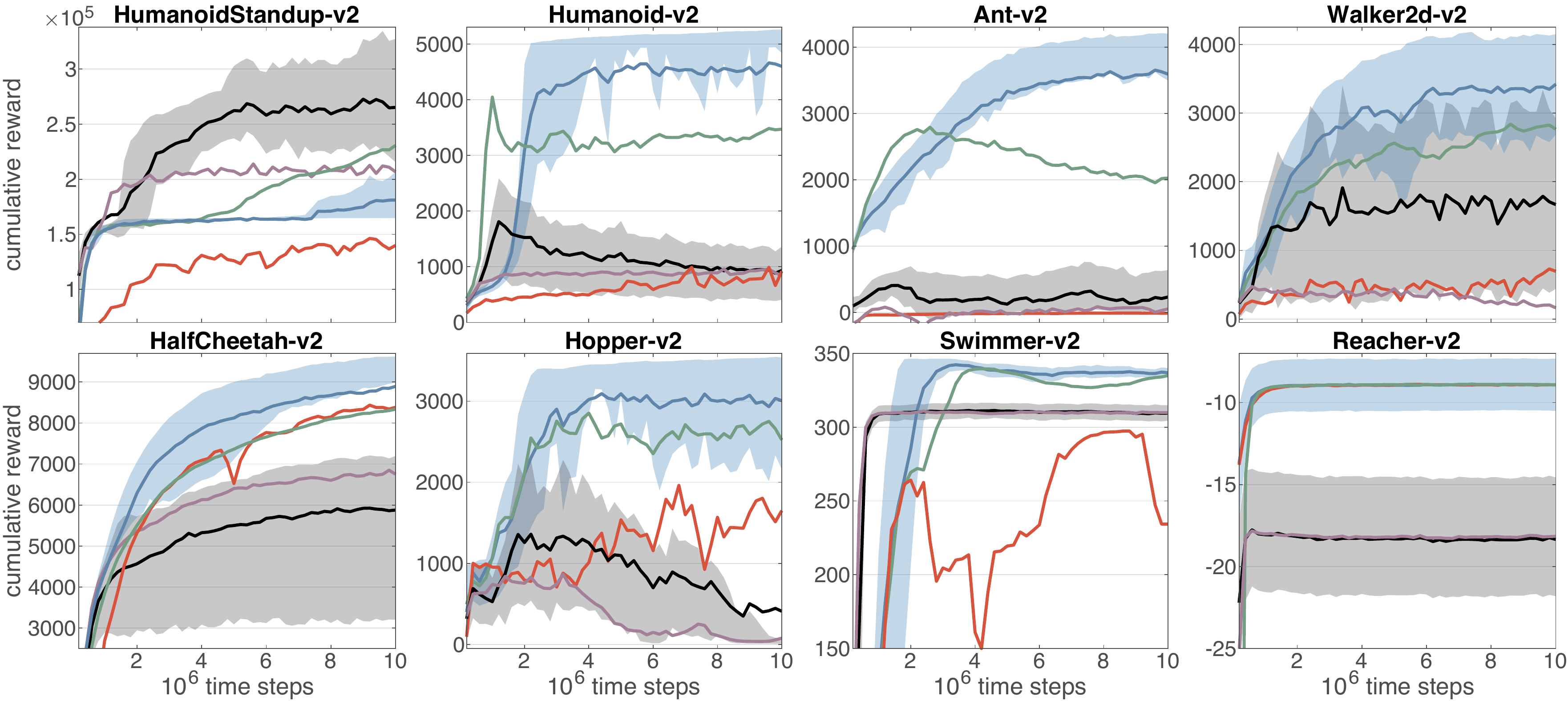}
	\caption{Cumulative rewards on OpenAI MuJoCo tasks for {\tt DDPG} (black line), {\tt DDPG} with rank-based {\tt PER} (purple line), {\tt DDPG} with {\tt ReF-ER} (blue), with {\tt ReF-ER-1} (red), and with {\tt ReF-ER-2} (green). Implementation details in App.~A.
	}\label{fig:benchdpg}
	\vspace{0.5cm}
	\includegraphics[width=\textwidth, trim={0 0 0 0}, clip]{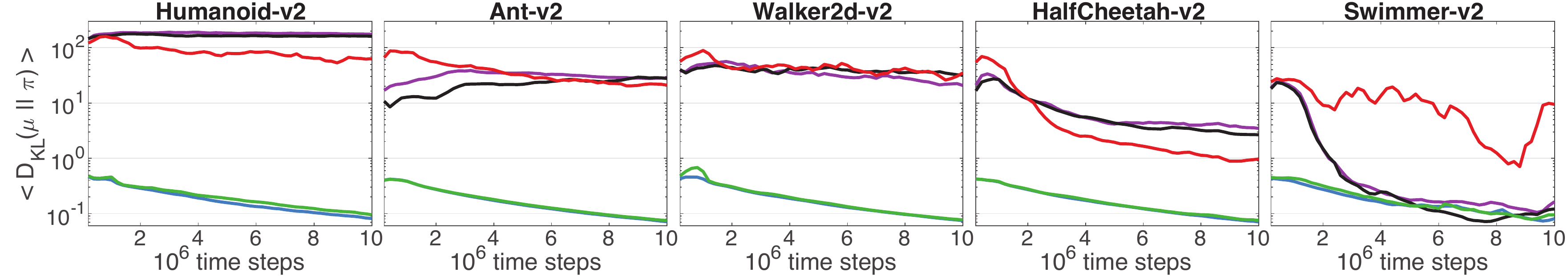}
\caption{Kullback-Leibler divergence $D_{KL}$ between $\pi^\nnw  {=} m^\nnw {+} \mathcal{N}(\mathbf{0}, \sigma^2 \mathbf{I})$ trained by {\tt DDPG} and the replayed behaviors. Same colors as above. Note: the average $D_{KL}$ for each algorithm is 0 at the beginning of training and is updated after every 1e5 time steps.}\label{fig:dklddpg}
\end{figure*}

The performance of {\tt DDPG} is sensitive to hyper-parameter (HP) tuning~\citep{henderson2017}. We find the critic's weight decay and temporally-correlated exploration noise to be necessary to stabilize {\tt DDPG} with ER and {\tt PER}. Without this tuning, the returns for {\tt DDPG} can fall to large negative values, especially in tasks that include the actuation cost in the reward (e.g. Ant). This is explained by the critic not having learned local maxima with respect to the action~\citep{silver2014}.
Fig.~\ref{fig:benchdpg} shows that replacing ER with {\tt ReF-ER} stabilizes {\tt DDPG} and greatly improves its performance, especially for tasks with complex dynamics (e.g. Humanoid and Ant). We note that with {\tt ReF-ER} we do not use temporally-correlated noise and that annealing $\eta$ worsened the instability of {\tt DDPG} with regular ER and {\tt PER}.

In Fig.~\ref{fig:dklddpg}, we report the average $D_{KL}(\mu_t || \pi^\nnw)$ as a measure of the RM's ``off-policyness''. With {\tt ReF-ER}, despite its reliance on approximating of the total number of far-policy samples $n_\text{far}$ in Eq.~\ref{eq:penl} from outdated importance weights, the $D_{KL}$ smoothly decreases during training due to the annealing process. This validates that Rule 2 of {\tt ReF-ER} achieves its intended goal with minimal computational overhead.
With regular ER, even after lowering $\eta$ by one order of magnitude from the original paper (we use $\eta{=}10^{{-}4}$ for the critic and $\eta{=}10^{{-}5}$ for the policy), $D_{KL}$ may span the entire action space. In fact, in many tasks the average $D_{KL}$ with ER is of similar order of magnitude as its maximum $2 d_A /\sigma^2$ ({\tt DDPG} by construction bounds $\textbf{m}^\nnw$ to the hyperbox $(-1, 1)^{d_A}$). For example, for $\sigma{=}0.2$, the maximum $D_{KL}$ is 850 for Humanoid and 300 for Walker and it oscillates during training around 100 and 50 respectively. This indicates that $\textbf{m}^\nnw$ swings between the extrema of the action space likely due to the critic not learning local maxima for $Q^\nnw$. Without policy constrains, {\tt DDPG} often finds only ``bang-bang'' control schemes, which explains why bounding the action space is necessary to ensure the stability of {\tt DDPG}.

\begin{figure*}[t] \centering
	\includegraphics[width=\textwidth, trim={0 0 0 0}, clip]{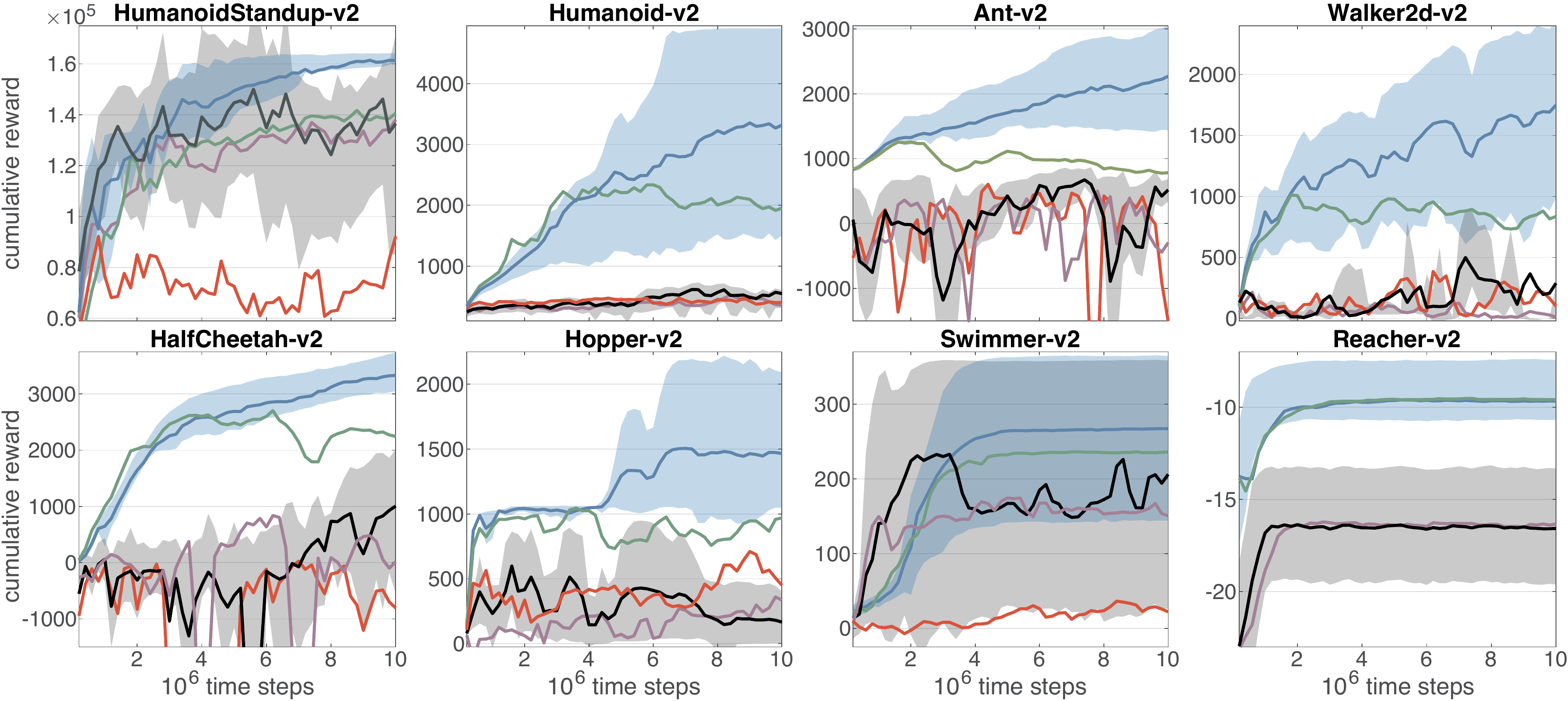}
	\caption{Cumulative rewards on OpenAI MuJoCo tasks for {\tt NAF} (black line), {\tt NAF} with rank-based {\tt PER} (purple line), {\tt NAF} with {\tt ReF-ER} (blue), with {\tt ReF-ER-1} (red), and with {\tt ReF-ER-2} (green). Implementation details in App.~A.
	}\label{fig:benchnaf}
\end{figure*}

When comparing the components of {\tt ReF-ER}, we note that relying on gradient clipping alone ({\tt ReF-ER-1}) does not produce good results. {\tt ReF-ER-1} may cause many zero-valued gradients, especially in high-dimensional tasks where even small changes to $\textbf{m}^\nnw$ may push $\rho_t$ outside of the near-policy region.
However, it's on these tasks that combining the two rules brings a measurable improvement in performance over {\tt ReF-ER-2}.
Training from only near-policy samples, provides the critic with multiple examples of trajectories that are possible with the current policy.
This focuses the representation capacity of the critic, enabling it to extrapolate the effect of a marginal change of action on the expected returns, and therefore increasing the accuracy of the DPG. Any misstep of the DPG is weighted with a penalization term that attracts the policy towards past behaviors. This allows time for the learner to gather experiences with the new policy, improve the value-network, and correct the misstep. This reasoning is almost diametrically opposed to that behind {\tt PER}, which generally obtains worse outcomes than regular ER. In {\tt PER} observations associated with larger TD errors are sampled more frequently. In the continuous-action setting, however, TD errors may be caused by actions that are farther from $\textbf{m}^\nnw$. Therefore, precisely estimating their value might not help the critic in yielding an accurate estimate of the DPG. 
The Swimmer and HumanoidStandup tasks highlight that ER is faster than {\tt ReF-ER} in finding bang–bang policies. The bounds imposed by {\tt DDPG} on $\textbf{m}^\nnw$ allow learning these behaviors without numerical instability and without finding local maxima of $Q^\nnw$. The methods we consider next learn unbounded policies. These methods do not require prior knowledge of optimal action bounds, but may not enjoy the same stability guarantees.

\subsection{Results for {\tt NAF}}\label{sec:nafret}
Figure~\ref{fig:benchnaf} shows how {\tt NAF} is affected by the choice of ER algorithm. While Q-learning based methods are thought to be less sensitive than PG-based methods to the dissimilarity between policy and stored behaviors owing to the bootstrapped Q-learning target, {\tt NAF} benefits from both rules of {\tt REF-ER}. Like for {\tt DDPG}, Rule 2 provides NAF with more near-policy samples to compute the off-policy estimators. Moreover, the performance of {\tt NAF} is more distinctly improved by combining Rule 1 and 2 of {\tt REF-ER} over using {\tt REF-ER-2}. This is because $Q^\pi$ is likely to be approximated well by the quadratic $Q^\nnw_\text{NAF}$ in a small neighborhood near its local maxima. When $Q^\nnw_\text{NAF}$ learns a poor fit of $Q^\pi$ (e.g. when the return landscape is multi-modal), {\tt NAF} may fail to choose good actions. 
Rule 1 clips the gradients from actions outside of this neighborhood and prevents large TD errors from disrupting the locally-accurate approximation $Q^\nnw_\text{NAF}$. This intuition is supported by observing that rank-based {\tt PER} (the better performing variant of {\tt PER} also in this case), often worsens the performance of {\tt NAF}. {\tt PER} aims at biasing sampling in favor of larger TD errors, which are more likely to be farther from $\mathbf{m}^\nnw(s)$, and their accurate prediction might not help the learner in fine-tuning the policy by improving a local approximation of the advantage. Lastly, $Q^\nnw_\text{NAF}$ is unbounded, therefore training from actions that are farther from $\mathbf{m}^\nnw$ increases the variance of the gradient estimates.

\begin{figure*}[t] \centering
	\includegraphics[width=\textwidth, trim={0 0 0 0}, clip]{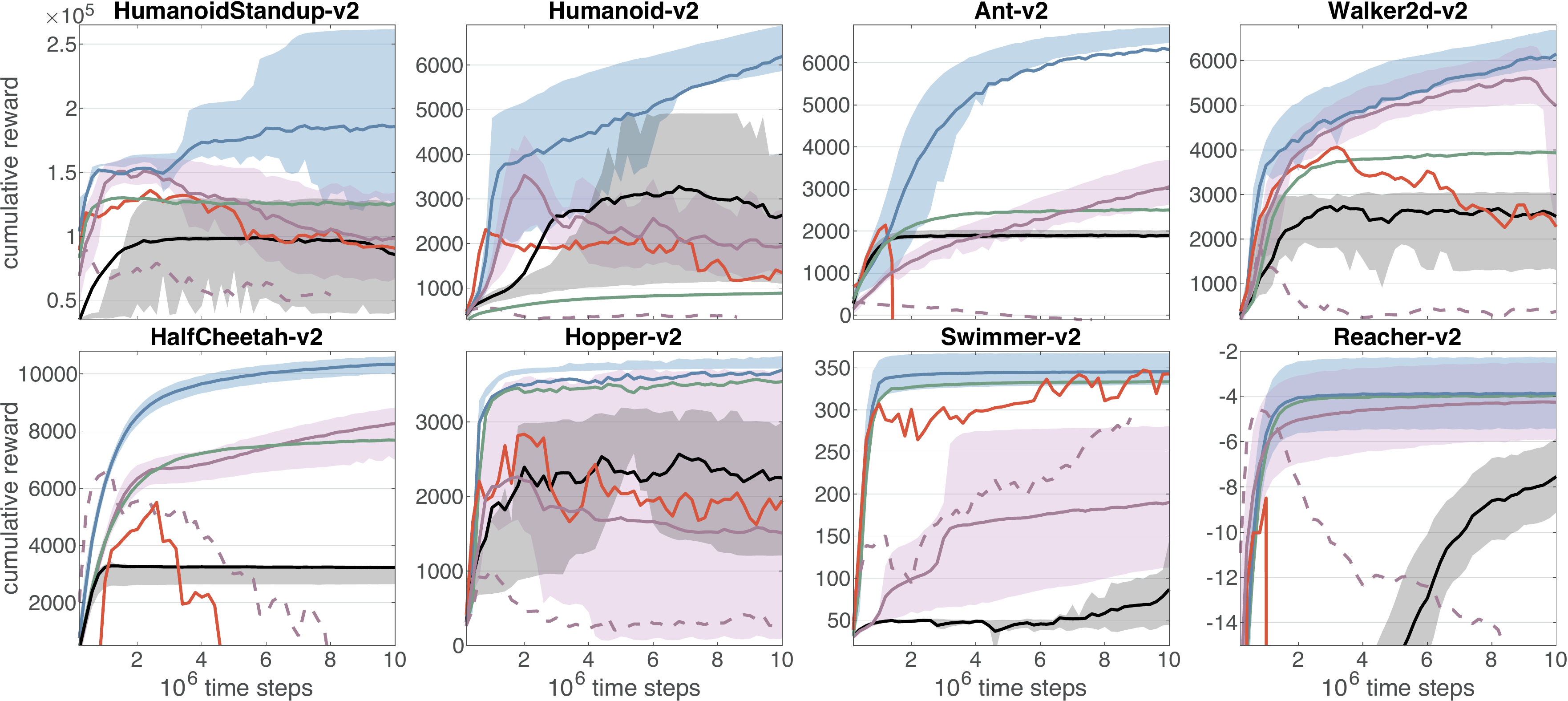}
	\caption{Average cumulative rewards  on MuJoCo OpenAI Gym tasks obtained by {\tt PPO} (black line), {\tt ACER} (purple dashed line for $\eta=10^{-4}$ and full line for $\eta=10^{-5}$) and {\tt V-RACER} with {\tt ReF-ER} (blue), {\tt ReF-ER-1} (red), {\tt ReF-ER-2} (green).}\label{fig:bench1}
	\vspace{0.5cm}
	\includegraphics[width=\textwidth, trim={0 0 0 0}, clip]{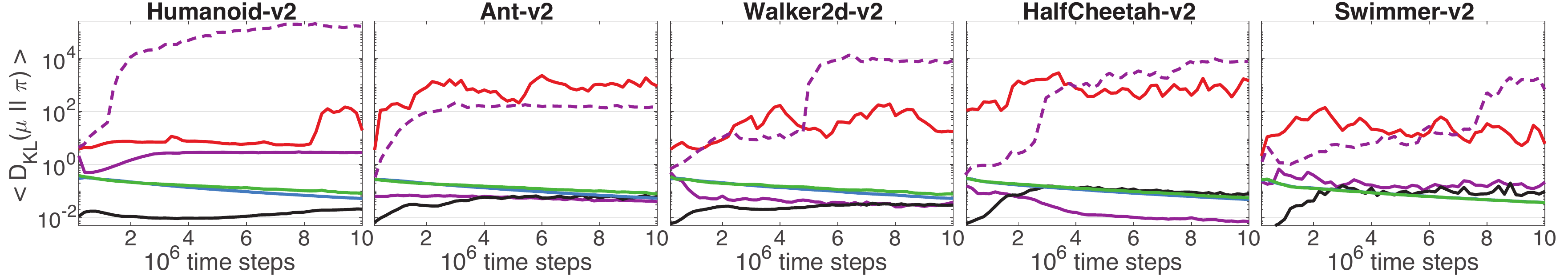}
	\caption{Kullback-Leibler divergence between $\pi^\nnw$ and the replayed behaviors obtained by the PG-based methods. Same colors as above.}\label{fig:dklacer}
\end{figure*}

\subsection{Results for {\tt V-RACER}}\label{sec:racret}
Here we compare {\tt V-RACER} to {\tt ACER} and to {\tt PPO}, an algorithm that owing to its simplicity and good performance on MuJoCo tasks is often used as baseline. For clarity, we omit from Fig.~\ref{fig:bench1} results from coupling {\tt V-RACER} with ER or {\tt PER}, which generally yield  similar or worse results than {\tt ReF-ER-1}. Without Rule 1 of {\tt ReF-ER}, {\tt V-RACER} has no means to deal with unbounded importance weights, which cause off-PG estimates to diverge and disrupt prior learning progress. In fact, also {\tt ReF-ER-2} is affected by unbounded $\rho_t$ because even small policy differences can cause $\rho_t$ to overflow if computed for actions at the tails of the policy. For this reason, the results of {\tt ReF-ER-2} are obtained by clipping all importance weights $\rho_t \leftarrow \min(\rho_t, 10^3)$.

Similarly to {\tt ReF-ER}, {\tt ACER}'s techniques (summarized in Sec. \ref{sec:related}) guard against the numerical instability of the off-PG. {\tt ACER} partly relies on constraining policy updates around a target-network with tuned learning and target-update rates. However, when using deep NN, small parameter updates do not guarantee small differences in the NN's outputs. Therefore, tuning the learning rates does not ensure similarity between $\pi^\nnw$ and RM behaviors. This can be observed in Fig.~\ref{fig:dklacer} from {\tt ACER}'s superlinear relation between policy changes and the NN's learning rate $\eta$. By lowering $\eta$ from $10^{-4}$ to $10^{-5}$, $D_{KL}(\mu_t~\|~\pi^\nnw)$ is reduced by multiple orders of magnitude (depending on the task). This corresponds to a large disparity in performance between the two choices of HP. For $\eta=10^{-4}$, as $D_{KL}(\mu_t~\|~\pi^\nnw)$ grows orders of magnitude more than with other algorithms, off-PG estimates become inaccurate, causing {\tt ACER} to be often outperformed by {\tt PPO}. These experiments, together with the analysis of {\tt DDPG}, illustrate the difficulty of controlling off-policyness in deep RL by enforcing slow parameter changes.

{\tt ReF-ER} aids off-policy PG methods in two ways. As discussed for {\tt DDPG} and {\tt NAF}, Rule 2 ensures a RM of valuable experiences for estimating on-policy quantities with a finite batch size. In fact, we observe from Fig.~\ref{fig:bench1} that {\tt ReF-ER-2} alone often matches or surpasses the performance of {\tt ACER}. Rule 1 prevents unbounded importance weights from increasing the variance of the PG and from increasing the amount of ``trace-cutting'' incurred by Retrace~\citep{munos2016}. Trace-cutting reduces the speed at which $Q^\text{ret}$ converges to the on-policy $Q^{\pi^\nnw}$ after each change to $\pi^\nnw$, and consequently affects the accuracy of the loss functions.
On the other hand, skipping far-policy samples without penalties or without extremely large batch sizes~\citep{openaidota} causes {\tt ReF-ER-1} to have many zero-valued gradients (reducing the effective batch size) and unreliable outcomes.

Annealing $c_\text{max}$ eventually provides {\tt V-RACER} with a RM of experiences that are almost as on-policy as those used by {\tt PPO}. In fact, while considered on-policy, {\tt PPO} alternates gathering a small RM (usually $2^{11}$ experiences) and performing few optimization steps on the samples. Fig.~\ref{fig:dklacer} shows the average $D_{KL}(\mu_t~\|~\pi^\nnw)$ converging to similar values for both methods. While a small RM may not contain enough diversity of samples for the learner to accurately estimate the gradients. The much larger RM of {\tt ReF-ER} (here $N=2^{18}$ samples), and possibly the continually-updated value targets, allow {\tt V-RACER} to obtain much higher returns. The Supplementary Material contains extended analysis of {\tt V-RACER}'s most relevant HP.
For many tasks presented here, {\tt V-RACER} combined with {\tt ReF-ER} outperforms the best result from {\tt DDPG} (Sec.~\ref{sec:ddpgret}), {\tt NAF} (Sec.~\ref{sec:nafret}), {\tt PPO} and {\tt ACER} and is competitive with the best published results, which to our knowledge were achieved by the on-policy algorithms Trust Region Policy Optimization~\citep{schulman2015a}, Policy Search with Natural Gradient~\citep{rajeswaran2017}, and Soft Actor-Critic~\citep{haarnoja2018}. 

\begin{figure}[t]
	\centering
	\includegraphics[width=0.65\linewidth, trim={0 0 0 0}, clip]{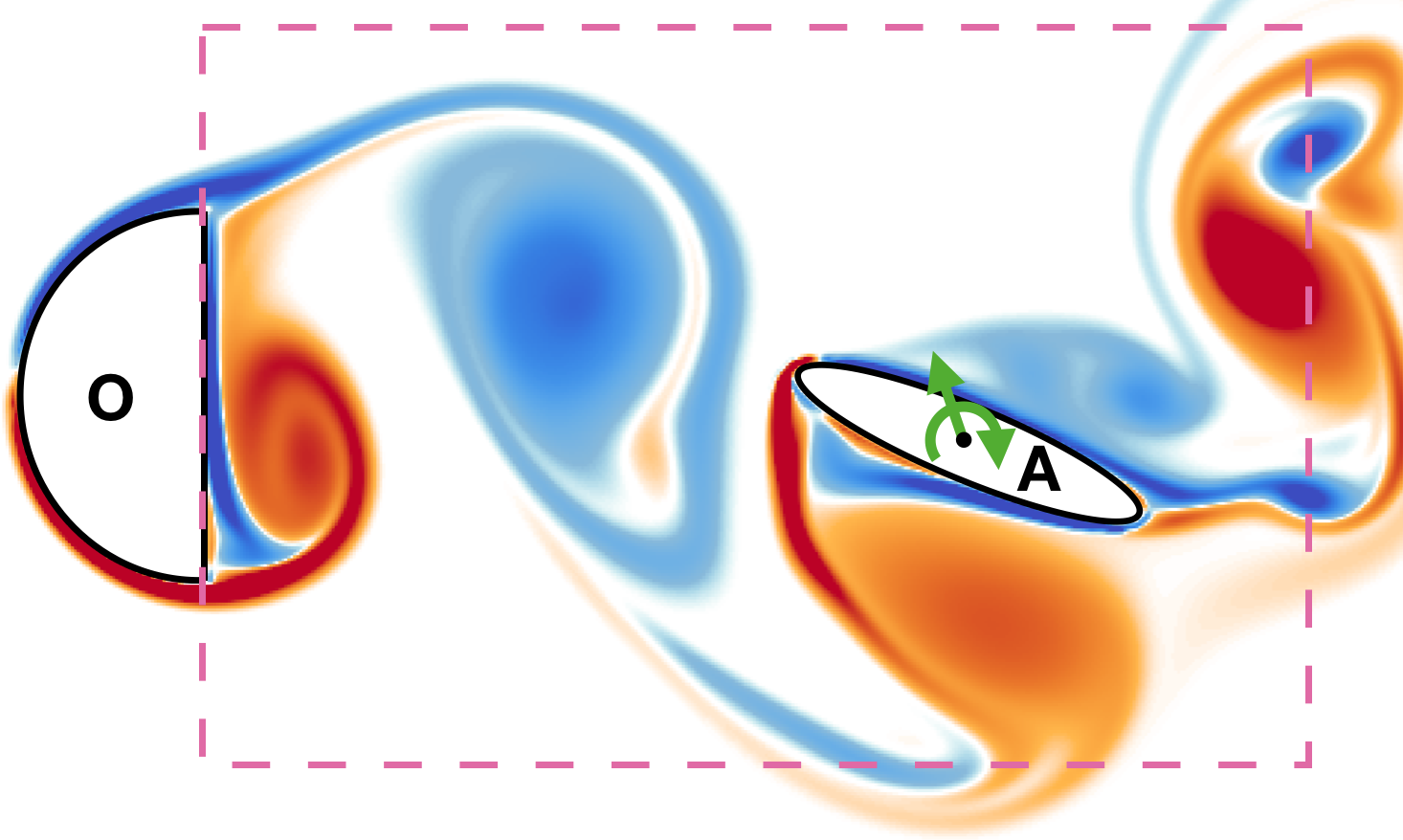} \hfill \includegraphics[width=0.34\linewidth]{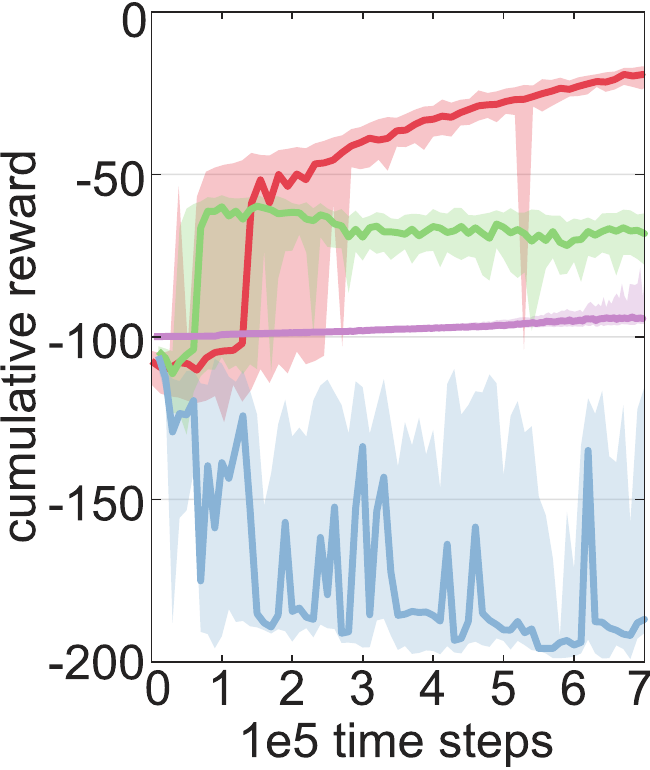}
	\caption{(left) Contours of the vorticity field (red and blue for anti- and clockwise rotation respectively) of the 2D flow control problem: the D-section cylinder is moving leftward, the agent is marked by A and by the highlighted control force and torque. (right) Returns obtained by {\tt V-RACER} (red), {\tt ACER} (purple), {\tt DDPG} with ER (blue), {\tt DDPG} with {\tt ReF-ER} (green).} \label{fig:smartcyl}
\end{figure}

\subsection{Results for a partially-observable flow control task}
The problems considered so far have been modeled by ordinary differential equations (ODE), with the agent having access to the entire state of the system. We now apply the considered methods to systems described by non-linear Partial Differential Equations (PDE), here the Navier Stokes  Equations (NSE) that govern continuum fluid flows. Such PDEs are used to describe many problems of scientific (e.g. turbulence, fish swimming) and industrial interests (e.g. wind farms, combustion engines). These problems pose two challenges: First, accurate simulations of PDEs may entail significant computational costs and large scale computing resources which exceed by several orders of magnitude what is required by ODEs. Second, the NSE are usually solved on spatial grids with millions or even trillions of degrees of freedom. It would be excessive to provide all that information to the agent, and therefore the state is generally measured by a finite number of sensors. Consequently, the assumption of Markovian dynamics at the core of most RL methods is voided. This may be remedied by using recurrent NN (RNN) for function approximation. In turn, RNNs add to the challenges of RL the increased complexity of properly training them. Here we consider the small 2D flow control problem of agent $A$, an elliptical body of major-axis $D$ and aspect ratio $0.2$, interacting with an unsteady wake. The wake is created by a D-section cylinder of diameter $D$ ($O$ in Fig.~\ref{fig:smartcyl}) moving at constant speed (one length $D$ per time-unit $\mathcal{T}$) at Reynolds number $D^2 / (\nu \mathcal{T}) {=} 400$. Agent $A$ performs one action per unit $\mathcal{T}$ by imposing a force and a torque on the flow $a_t {:=} \{f_X, f_Y, \tau\}$. The state $s_t{\in}\mathbb{R}^{14}$ contains $A$'s position, orientation and velocity relative to $O$ and has 4 flow-speed sensors located at $A$'s 4 vertices. The reward is $r_{t+1} {=} -\|a_t\|^2$. If $A$ exits the area denoted by a dashed line in Fig.~\ref{fig:smartcyl}, the terminal reward is $-100$ and the simulation restarts with random initial conditions. Otherwise, the maximum duration of the simulation is 400 actions. We attempt this problem with three differently-seeded runs of each method considered so far. Instead of maximizing the performances by HP tuning, we only substitute the MLPs used for function approximation with LSTM networks (2 layers of 32 cells with back-propagation window of 16 steps). 

If correctly navigated, drafting in the momentum released into the flow by the motion of $O$ allows $A$ to maintain its position with minimal actuation cost. 
Fig.~\ref{fig:smartcyl} shows that the optimal HP found for {\tt ACER} (small $\eta$) in the ODE tasks, together with the lack of feature-sharing between policy and value-networks and with the variance of the off-PG, cause the method to make little progress during training. {\tt DDPG} with ER incurs large actuation costs, while {\tt DDPG} with {\tt ReF-ER} is the fastest at learning to avoid the distance limits sketched in Fig.~\ref{fig:smartcyl}. In fact the critic quickly learns that $A$ needs to accelerate leftward to avoid being left behind, and the policy adopts the behavior rapidly due to the lower variance of the DPG~\citep{silver2014}. Eventually, the best performance is reached by {\tt V-RACER} with {\tt ReF-ER} (an animation of a trained policy is provided in the Supplementary Material). {\tt V-RACER} has the added benefit of having an unbounded action space and of feature-sharing: a single NN receives the combined feedback of $V^\nnw$ and $\pi^\nnw$ on how to shape its internal representation of the dynamics.

\section{Conclusion}
Many RL algorithms update a policy $\pi^\nnw$ from experiences collected with off-policy behaviors $\mu$. We present evidence that off-policy continuous-action deep RL methods benefit from actively maintaining similarity between policy and replay behaviors. 
We propose a novel ER algorithm ({\tt ReF-ER}) which consists of:
(1) Characterizing past behaviors either as ``near-policy" or ``far-policy" by the deviation from one of the importance weight $\rho{=} \pi^\nnw(a|s) / \mu (a|s)$ and computing gradients only from near-policy experiences. (2) Regulating the pace at which $\pi^\nnw$ is allowed to deviate from $\mu$ through penalty terms that reduce $D_{KL}(\mu || \pi^\nnw)$. This allows time for the learner to gather experiences with the new policy, improve the value estimators, and increase the accuracy of the next steps.
We analyze the two components of {\tt ReF-ER} and show their effects on continuous-action RL algorithms employing off-policy PG, deterministic PG ({\tt DDPG}) and Q-learning ({\tt NAF}). Moreover, we introduce {\tt V-RACER}, a novel algorithm based on the off-policy PG which emphasizes simplicity and computational efficiency. The combination of {\tt ReF-ER} and {\tt V-RACER} reliably yields performance that is competitive with the state-of-the-art. 

\subsubsection*{Acknowledgments}
We thank Siddhartha Verma for helpful discussions and feedback on this manuscript. This work was supported by European Research Council Advanced Investigator Award 341117. Computational resources were provided by Swiss National Supercomputing Centre (CSCS) Project s658 and s929.

{\small
\bibliographystyle{iclr2019_conference}
\bibliography{racer}
}

\ifprintsm

\appendix
\newpage

\twocolumn[
\icmltitle{Remember and Forget for Experience Replay\\
Supplementary Material }

\begin{icmlauthorlist}
	\icmlauthor{Guido Novati}{ethz}
	\icmlauthor{Petros Koumoutsakos}{ethz}
\end{icmlauthorlist}

\vskip 0.3in
]

\section{Implementation and network architecture details}\label{sec:deets}

We implemented all presented learning algorithms within {\tt smarties},\footnote{\url{https://github.com/cselab/smarties}}
our open source {\tt C++} RL framework, and optimized for high CPU-level efficiency through fine-grained multi-threading, strict control of cache-locality, and computation-communication overlap. On every step, we asynchronously obtain on-policy data by sampling the environment with $\pi$, which advances the index $t$ of observed time steps, and we compute updates by sampling from the Replay Memory (RM), which advances the index $k$ of gradient steps. During training, the ratio of time and update steps is equal to a constant $F = t/k$, usually set to 1. 
Upon completion of all tasks, we apply the gradient update and proceed to the next step. The pseudo-codes in App.~\ref{sec:pseudo} neglect parallelization details as they do not affect execution.

In order to evaluate all algorithms on equal footing, we use the same baseline network architecture for {\tt V-RACER}, {\tt DDPG} and {\tt NAF}, consisting of an MLP with two hidden layers of 128 units each. For the sake of computational efficiency, we employed Softsign activation functions. The weights of the hidden layers are initialized according to $\mathcal{U}\left[{-}6/\sqrt{f_i + f_o}, ~6/\sqrt{f_i + f_o} \right]$, where $f_i$ and $f_o$ are respectively the layer's fan-in and fan-out~\citep{glorot2010}.
The weights of the linear output layer are initialized from the distribution $\mathcal{U}\left[{-}0.1/\sqrt{f_i}, ~0.1/\sqrt{f_i} \right]$, such that the MLP has near-zero outputs at the beginning of training. When sampling the components of the action vectors, the policies are treated as truncated normal distributions with symmetric bounds at three standard deviations from the mean. Finally, we optimize the network weights with the Adam algorithm~\citep{kingma2014}.

\textbf{V-RACER} We note that the values of the diagonal covariance matrix are shared among all states and initialized to $\boldsymbol{\Sigma} {=} 0.2\mathbf{I}$. To ensure that $\boldsymbol{\Sigma}$ is positive definite, the respective NN outputs are mapped onto $\mathbb{R}^+$ by a Softplus rectifier. We set the discount factor $\gamma{=}0.995$, {\tt ReF-ER} parameters $C{=}4$, $A{=}5{\cdot}10^{-7}$ and $D{=}0.1$, and the RM contains $2^{18}$ samples. We perform one gradient step per environment time step, with mini-batch size $B{=}256$ and learning rate $\eta {=} 10^{-4}$.

\textbf{DDPG} We use the common MLP architecture for each network. The output of the policy-network is mapped onto the bounded interval $[-1, 1]^{d_A}$ with an hyperbolic tangent function. We set the learning rate for the policy-network to $1\cdot10^{-5}$ and that of the value-network to $1\cdot 10^{-4}$ with L2 weight decay coefficient of $1\cdot 10^{-4}$. The RM is set to contain $N{=}2^{18}$ observations and we follow~\citet{henderson2017} for the remaining hyper-parameters: mini-batches of $B{=}128$ samples, $\gamma{=}0.995$, soft target-network update coefficient $0.01$. We note that while DDPG is the only algorithm employing two networks, choosing half the batch-size as {\tt V-RACER} and {\tt NAF} makes the compute cost roughly equal among the three methods. Finally, when using {\tt ReF-ER} we add exploratory Gaussian noise  to the deterministic policy: $\pi^{\tt w'}  {=} \mathbf{m}^{\tt w'} {+}\mathcal{N}(\mathbf{0}, \sigma^2 \mathbf{I})$ with $\sigma {=} 0.2$. When performing regular ER or {\tt PER} we sample the exploratory noise from an Ornstein–Uhlenbeck process with $\sigma {=} 0.2$ and $\theta {=} 0.15$. 

\textbf{NAF} We use the same baseline MLP architecture and learning rate $\eta=10^{-4}$, batch-size $B=256$, discount $\gamma=0.995$, RM size $N=2^{18}$, and soft target-network update coefficient $0.01$. Gaussian noise is added to the deterministic policy $\pi^{\tt w'}  {=} \mathbf{m}^{\tt w'} {+}\mathcal{N}(\mathbf{0}, \sigma^2 \mathbf{I})$ with $\sigma {=} 0.2$.

\textbf{PPO} We tuned the hyper-parameters as~\citet{henderson2017}: $\gamma{=}0.995$, GAE $\lambda{=}0.97$, policy gradient clipping at $\Delta \rho_t {=} 0.2$, and we alternate performing 2048 environment steps and 10 optimizer epochs with batch-size 64 on the obtained data. Both the policy- and the value-network are 2-layer MLPs with 64 units per layer. We further improved results by having separate learning rates ($10^{-4}$ for the policy and $3\cdot 10^{-4}$ for the critic) with the same annealing as used in the other experiments.

\textbf{ACER} We kept most hyper-parameters as described in the original paper~\citep{wang2016}: the TBC clipping parameter is $c=5$, the trust-region update parameter is $\delta=1$, and five samples of the advantage-network are used to compute $A^\nnw$ estimates under $\pi$. We use a RM of $1e5$ samples, each gradient is computed from 24 uniformly sampled episodes, and we perform one gradient step per environment step. Because here learning is not from pixels, each network (value, advantage, and policy) is an MLP with 2 layers and 128 units per layer. Accordingly, we reduced the soft target-network update coefficient ($\alpha = 0.001$) and the learning rates for the advantage-network ($\eta{=}10^{-4}$), value-network ($\eta{=}10^{-4}$) and for the policy-network ($\eta{=}10^{-5}$).

\section{State, action and reward preprocessing}\label{sec:scaling}
Several authors have employed state~\citep{henderson2017} and reward~\citep{duan2016}~\citep{gu2017} rescaling to improve the learning results. 
For example, the stability of {\tt DDPG} is affected by the L2 weight decay of the value-network. Depending on the numerical values of the distribution of rewards provided by the environment and the choice of weight decay coefficient, the L2 penalization can be either negligible or dominate the Bellman error.
Similarly, the distribution of values describing the state variables can increase the challenge of learning by gradient descent.

We partially address these issues by rescaling both rewards and state vectors depending on the the experiences contained in the RM. At the beginning of training we prepare the RM by collecting $N_\text{start}$ observations and then we compute:
\begin{eqnarray}
\mu_s =& \frac{1}{n_\text{obs}}  \sum_{t=0}^{n_\text{obs}} s_t\\
\sigma_s =&  \sqrt{\frac{1}{n_\text{obs}}  \sum_{t=0}^{n_\text{obs}} \left( s_{t} - \mu_s \right)^2}
\end{eqnarray}
Throughout training, $\mu_s$ and $\sigma_s$ are used to standardize all state vectors $\hat{s}_t = (s_t - \mu_s)/(\sigma_s + \epsilon)$ before feeding them to the NN approximators.
Moreover, every 1000 steps, chosen as the smallest power of 10 that doesn't affect the run time, we loop over the $n_\text{obs}$ samples stored in the RM to compute:
\begin{equation}
\sigma_r \leftarrow \sqrt{\frac{1}{n_\text{obs}} \sum_{t=0}^{n_\text{obs}} \left( r_{t+1} \right)^2}
\end{equation}
This value is used to scale the rewards $\hat{r}_{t} = r_t / (\sigma_r + \epsilon)$ used by the Q-learning target and the Retrace algorithm. We use $\epsilon = 10^{-7}$ to ensure numerical stability.

The actions sampled by the learner may need to be rescaled or bounded to some interval depending on the environment. For the OpenAI Gym tasks this amounts to a linear scaling $a' {=} a ~ (\verb!upper_value! - \verb!lower_value!)/2$, where the values specified by the Gym  library are $\pm 0.4$ for Humanoid tasks, $\pm8$ for Pendulum tasks, and  $\pm1$ for all others.

\section{Pseudo-codes}\label{sec:pseudo}
\begin{algorithm}[t]
	\caption{Serial description of the master algorithm.}
	\label{alg:master}
	\begin{algorithmic}
		\STATE $t = 0$, $k=0$,
		\STATE Initialize an empty RM, network weights $\nnw$, and {\tt Adam}'s~\citep{kingma2014} moments.
		\WHILE{$n_\text{obs} < N_\text{start}$}
		\STATE Advance the environment according algorithm~\ref{alg:expl}.
		\ENDWHILE
		\STATE Compute the initial statistics used to standardize the state vectors (App.~\ref{sec:scaling}).
		\STATE Compute the initial statistics used to rescale the rewards (App.~\ref{sec:scaling}).
		\WHILE{$t < T_\text{max}$}
		\WHILE{$t < F \cdot k$}
		\STATE  Advance the environment according to algorithm~\ref{alg:expl}.
		\WHILE{ $n_\text{obs} > N_\text{start}$ }
		\STATE Remove an episode from RM (first in first out).
		\ENDWHILE
		\STATE  $t \leftarrow t + 1$
		\ENDWHILE
		\STATE Sample $B$ time steps from the RM to compute a gradient estimate (e.g. for {\tt V-RACER} with algorithm~\ref{alg:racerlearn}).
		\STATE Perform the gradient step with the Adam algorithm.
		\STATE If applicable, update the {\tt ReF-ER} penalization coefficient $\beta$.
		\IF{$\text{modulo}(k, 1000)$ is 0 }
		\STATE Update the statistics used to rescale the rewards (App.~\ref{sec:scaling}).
		\ENDIF
		\STATE  $k \leftarrow k + 1$
		\ENDWHILE
	\end{algorithmic}
\end{algorithm}
Remarks on algorithm~\ref{alg:master}: 1) It describes the general structure of the ER-based off-policy RL algorithms implemented for this work (i.e. {\tt V-RACER}, {\tt DDPG}, and {\tt NAF}). 2) This algorithm can be adapted to conventional ER, {\tt PER} (by modifying the sampling algorithm to compute the gradient estimates), or {\tt ReF-ER} (by following Sec.~\ref{sec:ER})). 3) The algorithm requires 3 hyper-parameters: the ratio of time step to gradient steps $F$ (usually set to 1 as in DDPG), the maximal size of the RM $N$, and the minimal size of the RM before we begin gradient updates $N_\text{start}$. 

\begin{algorithm}[t]
	\caption{Environment sampling}
	\label{alg:expl}
	\begin{algorithmic}
		\STATE Observe $s_{t}$ and $r_{t}$.
		\IF{$s_{t}$ concludes an episode}
		\STATE Store data for $t$ into the RM: $\{s_t, r_t, V^\nnw (s_t)\}$
		\STATE Compute and store $Q^\text{ret}$ for all steps of the episode
		\ELSE
		\STATE Sample the current policy $a_t \sim \pi^\nnw (a | s_t) = \mu_t$
		\STATE Store data for $t$ into the RM: $\{s_t, r_t, a_t, \mu_t, V^\nnw (s_t)\}$
		\STATE Advance the environment by performing $a_t$
		\ENDIF
	\end{algorithmic}
\end{algorithm}
\begin{algorithm}[!t]
\caption{{\tt V-RACER}'s gradient update}
\label{alg:racerlearn}
\begin{algorithmic}
	\FOR{mini-batch sample $i=0$ to $B$}
	\STATE Fetch all relevant information: $s_{t_i}$, $a_{t_i}$, $V^\text{tbc}_{t_i}$, and $\mu_{t_i} = \{ \mathbf{m}_{t_i}, \boldsymbol{\Sigma}_{t_i}\}$.
	\STATE Call the approximator to compute $\pi^\nnw$ and $V^\nnw (s_{t_i})$
	\STATE Update $\rho_{t_i} =  \pi^\nnw (a_{t_i} | s_{t_i} )/\mu_{t_i}(a_{t_i} | s_{t_i})$
	\STATE Update $V^\text{tbc}$ for all prior steps in ${t_i}$'s episode with $V^\nnw (s_{t_i})$ and $\rho_{t_i}$
	\IF{$1/c_{\text{max}} <\rho_{t_i} < c_{\text{max}}$}
		\STATE Compute $\hat{g}_{t_i}(\nnw)$ according to Sec.~\ref{sec:pre}
	\ELSE
		\STATE $\hat{g}_{t_i}(\nnw) = \mathbf{0}$
	\ENDIF
	\STATE {\tt ReF-ER} penalization: $\hat{g}^\text{ReF-ER}_{t_i}(\nnw)  = \beta \hat{g}_{t_i}(\nnw) - (1{-}\beta) \nabla D_{\text{KL}} [\mu_{t_i}(\cdot | s_{t_i}) || \pi^\nnw (\cdot | s_{t_i})]$
	\ENDFOR
	\STATE  Accumulate the gradient estimate over the mini-batch $\frac{1}{B}\sum_{i=0}^B \hat{g}^\text{ReF-ER}_{t_i}(\nnw)$
\end{algorithmic}
\end{algorithm}
\begin{algorithm}[!t]
	\caption{{\tt DDPG}'s gradient update with {\tt ReF-ER}}
	\label{alg:ddpglearn}
	\begin{algorithmic}
		\FOR{mini-batch sample $i=0$ to $B$}
		\STATE Fetch all relevant information: $s_{t_i}$, $a_{t_i}$, and $\mu_{t_i} = \{ \mathbf{m}_{t_i}, \boldsymbol{\Sigma}_{t_i}\}$.
		\STATE The policy-network computes $\mathbf{m}^\nnw(s_{t_i})$ and the value-network computes $Q^{\nnw'} (s_{t_i}, a_{t_i})$. 
		\STATE Define a stochastic policy with Gaussian exploration noise: $\pi^\nnw(a~|~s_{t_i}) = \mathbf{m}^\nnw (s_{t_i})+ \mathcal{N} $
		\STATE Update $\rho_{t_i} =  \pi^\nnw (a_{t_i} | s_{t_i} )/\mu_{t_i}(a_{t_i} | s_{t_i})$
		\IF{$1/c_{\text{max}} <\rho_{t_i} < c_{\text{max}}$}
		\STATE Compute the policy at $t_i {+} 1$ with the target-network: $\mathbf{m}^{\tilde{\nnw}}(s_{t_i {+}1})$
		\STATE Compute the Q-learning target: $\hat{q}_{t_i} = r_{t_i+1} + \gamma Q^{\tilde{{\tt w}}'}\left(s_{t_i {+}1},~ \mathbf{m}^{\tilde{\nnw}}(s_{t_i {+}1}) \right) $
		\STATE The gradient $g^{Q}_{t_i}(\nnw')$ of the value-network minimizes the squared distance from $\hat{q}_{t_i}$.
		\STATE The gradient $g^\text{DPG}_{t_i}(\nnw)$ of the policy-network is the deterministic PG (Eq.~\ref{eq.dpg}).
		\ELSE
		\STATE $g^{Q}_{t_i}(\nnw') = \mathbf{0}$, $g^\text{DPG}_{t_i}(\nnw) =  \mathbf{0}$
		\ENDIF
		\STATE {\tt ReF-ER} penalization: $\hat{g}^\text{ReF-ER}_{t_i}(\nnw)  = \beta g^\text{DPG}_{t_i}(\nnw) - (1{-}\beta) \nabla D_{\text{KL}} [\mu_{t_i}(\cdot | s_{t_i}) || \pi^\nnw (\cdot | s_{t_i})]$
		\ENDFOR
		\STATE  Accumulate the gradient estimates over the mini-batch for both networks.
		\STATE Update the target policy- ($\tilde{\nnw} \leftarrow (1{-}\alpha)\tilde{\nnw} + \alpha \nnw$) and target value-networks ($\tilde{\nnw}' \leftarrow (1{-}\alpha)\tilde{\nnw}' + \alpha \nnw'$).
	\end{algorithmic}
\end{algorithm}

Remarks on algorithm~\ref{alg:expl}: 1) The reward for an episode's initial state, before having performed any action, is zero by definition. 2) The value $V^\nnw(s_t)$ for the last state of an episode is computed if the episode has been truncated due the task's time limits or is set to zero if $s_t$ is a terminal state. 3) Each time step we use the learner's updated policy-network and we store $\mu_t = \{\mathbf{m}(s_t), \boldsymbol{\Sigma}(s_t) \}$.

Remarks on algorithm~\ref{alg:racerlearn}: 1) In order to compute the gradients we rely on value estimates $V^\text{tbc}_{t_i}$ that were computed when subsequent time steps in $t_i$'s episode were previously drawn by ER. Not having to compute the quantities $V^\nnw$, and $\rho$ for all following steps comes with clear computational efficiency benefits, at the risk of employing an incorrect estimate for $Q^\text{ret}_{t_i}$. In practice, we find that the Retrace values incur only minor changes between updates (even when large RM sizes decrease the frequency of updates to the Retrace estimator) and that relying on previous estimates has no evident effect on performance. This could be attributed to the gradual policy changes enforced by {\tt ReF-ER}. 2) With a little abuse of the notation, with $\pi$ (or $\mu$) we denote the statistics (mean, covariance) of the multivariate normal policy, with $\pi(a | s )$ we denote the probability of performing action $a$ given state $s$, and with $\pi (\cdot | s)$ we denote the probability density function over actions given state $s$.

Remarks on algorithm~\ref{alg:ddpglearn}: 1) It assumes that weights and Adam are initialized for both policy-network and value-network. 2) The ``target'' weights are initialized as identical to the ``trained'' weights. 3) For the sake of brevity, we omit the algorithm for {\tt NAF}, whose structure would be very similar to this one. The key difference is that {\tt NAF} employs only one network and all the gradients are computed from the Q-learning target.

\section{Flow control simulation and parallelization}\label{sec:flowcontrol}
The Navier-Stokes equations are solved with our in-house 2D flow solver, parallelized with CUDA and OpenMP. We write the NSE with discrete explicit time integration as:
\begin{equation}
u^{k+1} = u^k + \delta t \left[ - (u^k \cdot \nabla) u^k + \nu \Delta u^k - \nabla P^k + F_s^k\right] \label{eq:ns}
\end{equation}
Here $u$ is the velocity field, $P$ is the pressure computed with the projection method by solving the Poisson equation $\nabla P = - \frac{1}{\delta t} \nabla {\cdot} u^k$~\citep{chorin1967}, and $F_s$ is the penalization force introduced by Brinkman penalization. The Brinkman penalization method~\citep{angot1999} enforces the no-slip and no-flow-through boundary conditions at the surface of the solid bodies by extending the NSE inside the body and introducing a forcing term. 
Furthermore, we assumed incompressibility and no gravitational effects with $\rho {:=} 1$. The simulations are performed on a grid of extent $8D$ by $4D$, uniform spacing $h{=}2^{-10}D$ and Neumann boundary conditions. The time step is limited by the condition $\delta t {=} 0.1~h / \max(\|u\|_\infty)$. Due to the computational cost of the simulations, we deploy 24 parallel agents, all sending states and receiving actions from a central learner. To preserve the ratio $F = t/k$ we consider $t$ to be the global number of simulated time-steps received by the learner.

\begin{figure*}[!t] \centering
	\includegraphics[width=\textwidth]{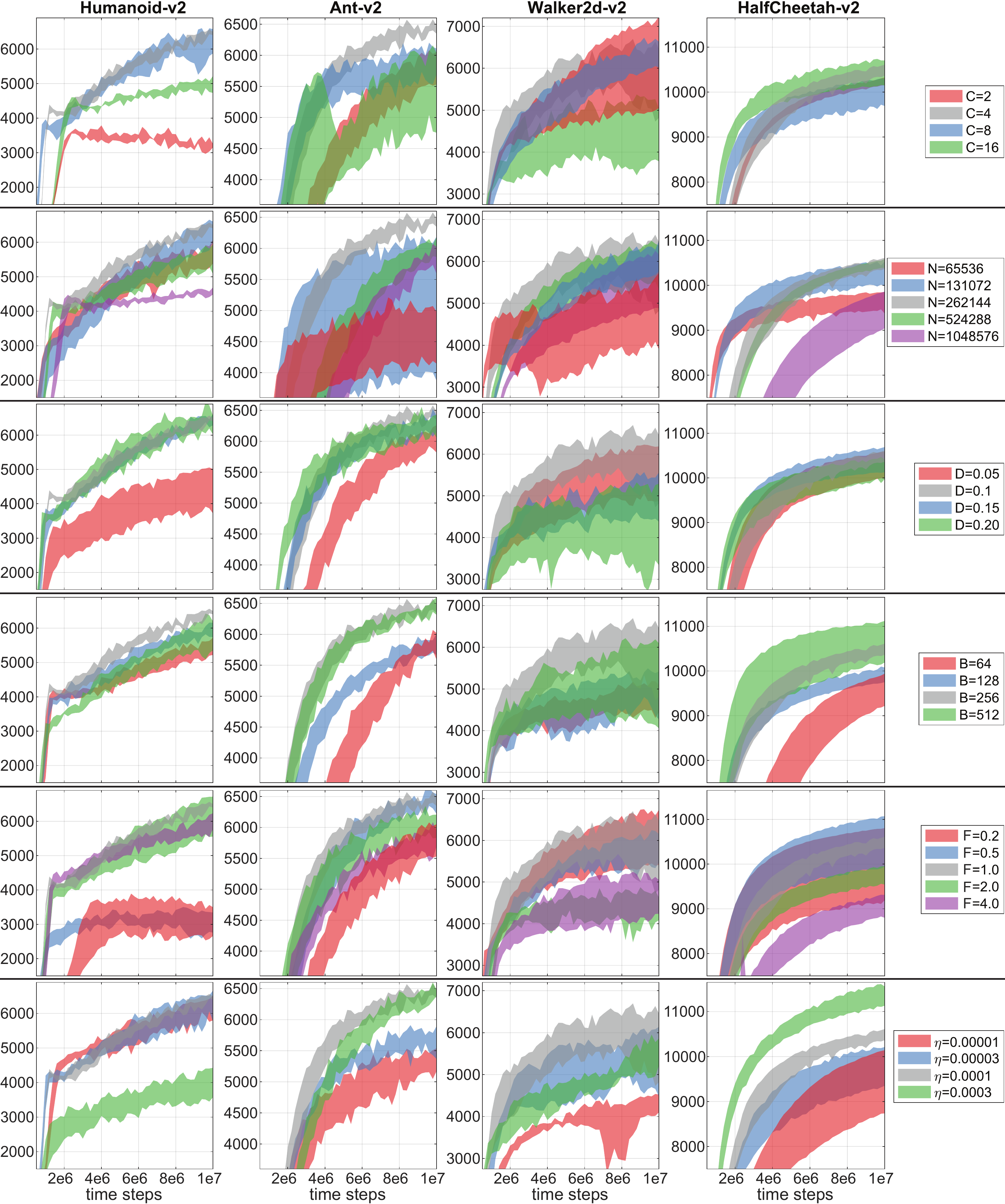}
	\caption{20\textsuperscript{th} and 80\textsuperscript{th} percentiles of the mean cumulative rewards over 5  training runs on a subset of OpenAI Gym tasks obtained with {\tt V-RACER}. In each row we vary one HP: the {\tt ReF-ER} parameters $C$ and $D$, the RM size $N$, mini-batch size $B$, number of time steps per gradient step $F$, and learning rate $\eta$.}\label{fig:HPtrain}
\end{figure*}
\begin{figure*}[t] \centering
	\includegraphics[width=\textwidth]{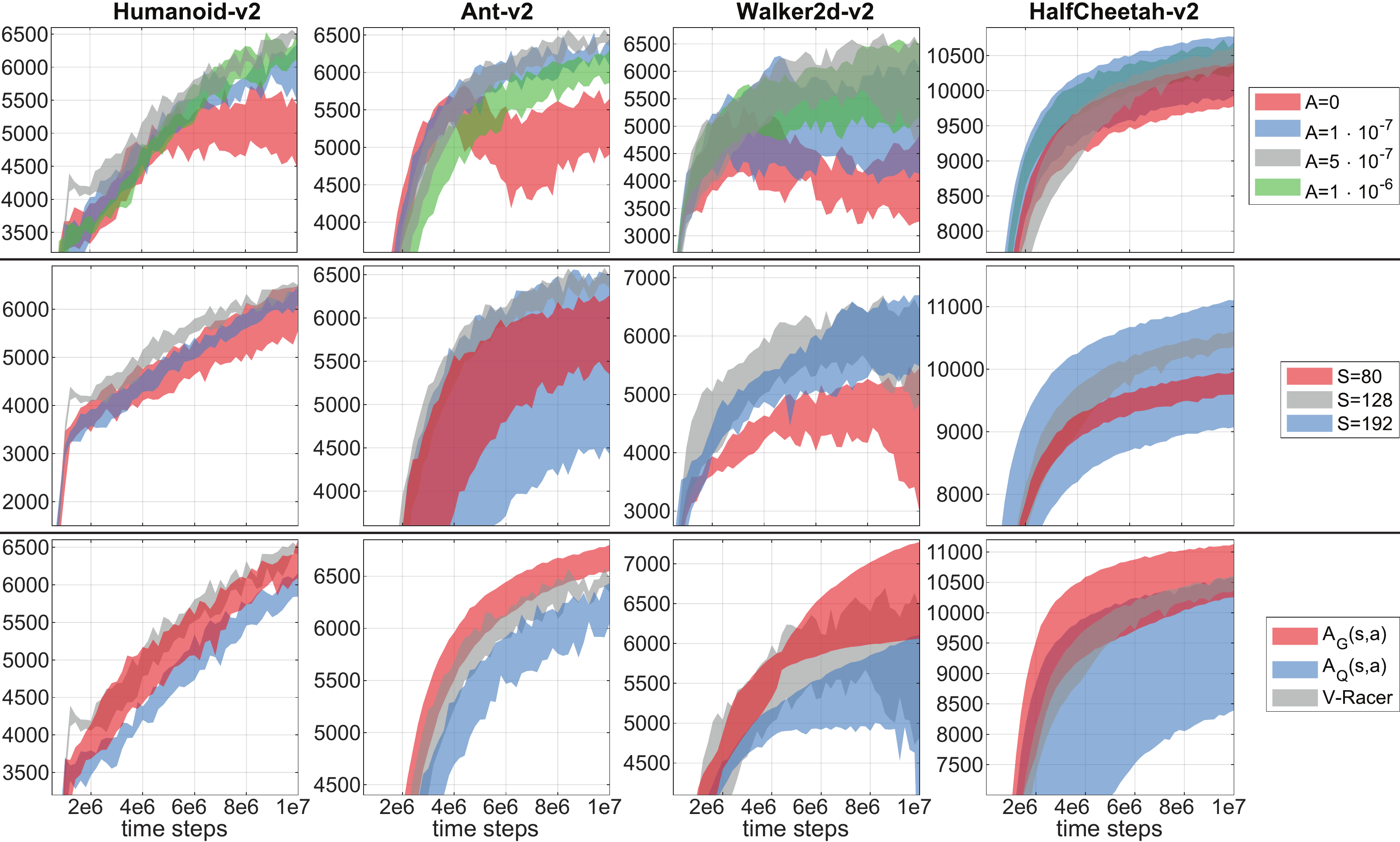}
	\caption{20\textsuperscript{th} and 80\textsuperscript{th} percentiles of the mean cumulative rewards over 5  training runs on a subset of OpenAI Gym tasks obtained with {\tt V-RACER} by varying the annealing schedule parameter $A$, the number $S$ of units in each of the two MLP hidden layers, and by extending the architecture with a parameterized action advantage $A^\nnw(s,a)$ as described in Sec.~\ref{sec:hyp}.}\label{fig:HParch}
\end{figure*}

\section{Sensitivity to hyper-parameters} \label{sec:hyp}
We report in Fig.~\ref{fig:HPtrain} and Fig.~\ref{fig:HParch} an extensive analysis of  {\tt V-RACER}'s robustness to the most relevant hyper-parameters (HP). The figures in the main text show the 20\textsuperscript{th} and 80\textsuperscript{th} percentiles of all cumulative rewards obtained over 5 training runs, binned in intervals of $2\cdot10^5$ time steps. Here we show the 20\textsuperscript{th} and 80\textsuperscript{th} percentiles of the mean cumulative rewards obtained over 5 training runs. This metric yields tighter uncertainty bounds and allows to more clearly distinguish the minor effects of HP changes.

The two HP that characterize the performance of {\tt ReF-ER} are the RM size $N$ and the importance sampling clipping parameter $c_\text{max} = 1 +C/(1+A\cdot t)$, where A is the annealing parameter discussed in Sec.~\ref{sec:ER}.
Both $C$ and $N$ determine the pace of policy changes allowed by {\tt ReF-ER}. Specifically, the penalty terms imposed by {\tt ReF-ER} increase for low values of $C$, because the trust region around replayed behaviors is tightened. On the other hand, high values of $C$ increase the variance of the gradients and may reduce the accuracy of the return-based value estimators by inducing trace-cutting~\citep{munos2016}. The penalty terms imposed by {\tt ReF-ER} also increase for large RM sizes $N$, because the RM is composed of episodes obtained with increasingly older behaviors which are all kept near-policy. Conversely, gradients computed from a small RM may be inaccurate because the environment's dynamics are not sufficiently covered by the training data.
These arguments are supported by the results in the first two rows of Fig.~\ref{fig:HPtrain}. Moreover, we observe that ``stable'' tasks, where the agent's success is less predicated on avoiding mistakes that would cause it to trip (e.g. HalfCheetah), are more tolerant to high values of $C$.

The tolerance $D$ for far-policy samples in the RM has a similar effect as $C$: low values of $D$ tend to delay learning while high values reduce the fraction of the RM that is used to compute updates and may decrease the accuracy of gradient estimates. The training performance can benefit from minor improvements by increasing the mini-batch size $B$, while the optimal learning rate $\eta$ is task-specific. From the first two rows of Fig.~\ref{fig:HParch} we observe that both the annealing schedule parameter $A$ and the number $S$ of units per layer of the MLP architecture have minor effects on performance. The annealing parameter $A$ allows the learner to fine-tune the policy parameters with updates computed from almost on-policy data at the later stages of training. We also note that wider networks may exhibit higher variance of outcomes. 

More uncertain is the effect of the number $F$ of environment time steps per learner's gradient step. Intuitively, increasing $F$ could either cause a rightward shift of the expected returns curve, because the learner computes fewer updates for the same budget of observations, or it could improve returns by providing more on-policy samples, which decreases {\tt ReF-ER}'s penalty terms and may increase the accuracy of the estimators. In practice the effect of $F$ is task-dependent. Problems with more complex dynamics, or higher dimensionality (e.g. Humanoid), seem to benefit from refreshing the RM more frequently with newer experiences (higher $F$), while simpler tasks can be learned more rapidly by performing more gradient steps per time step.

We considered extending {\tt V-RACER}'s architecture by adding a closed-form parameterization for the action advantage. Rather than having a separate MLP with inputs ($s, a$) to parameterize $Q^\nnw$ (as in {\tt ACER} or {\tt DDPG}), whose expected value under the policy would be computationally demanding to compute, we employ closed-form equations for $A^\nnw$ inspired by {\tt NAF}~\citep{gu2016}.
The network outputs the coefficients of a concave function $f^\nnw(s, a)$ which is chosen such that its maximum coincides with the mean of the policy $\mathbf{m}(s)$, and such that it is possible to derive analytical expectations for $a\sim\pi^\nnw$. Here we consider two options for the parameterized advantage. First, the quadratic form employed by {\tt NAF}~\citep{gu2016}:
\begin{equation}
f^\nnw_\text{Q}(s,a) = {-}\frac{1}{2} \left[ a {-} \mathbf{m}(s)\right]^\intercal \mathbf{L}_\text{Q}(s) \mathbf{L}_\text{Q}^\intercal(s) \left[ a {-} \mathbf{m}(s)\right]
\end{equation}
From $f^\nnw_\text{Q}$, the advantage is uniquely defined for any action $a$ as $A_\text{Q}^\nnw(s, a)  \vcentcolon=~ f_\text{Q}^\nnw(s, a) {-} \mathbb{E}_{a'\sim\pi}\left[f_\text{Q}^\nnw(s, a')\right]$. Therefore, like the exact on-policy advantage $A^\pi$, $A^\nnw$ has by design expectation zero under the policy. The expectation can be computed as~\citep{petersen2008}:
\begin{equation}
\mathbb{E}_{a'\sim\pi}\left[f^\nnw_\text{Q}(s, a')\right] = \text{Tr}\left[\mathbf{L}_\text{Q}(s) \mathbf{L}_\text{Q}^\intercal(s) \Sigma(s) \right]
\end{equation}
Here Tr denotes the trace of a matrix. 
Second we consider the asymmetric Gaussian parameterization:
\begin{equation}
f^\nnw_\text{G}(s,a) = K(s)~e^{\left[ -\frac{1}{2} \mathbf{a}_+^\intercal ~\mathbf{L}\inv_+(s)~ \mathbf{a}_+  -\frac{1}{2} \mathbf{a}_-^\intercal ~\mathbf{L}\inv_-(s) ~\mathbf{a}_-\right] }\label{eq:adv}
\end{equation}
Here $ \mathbf{a}_{-} {=} \min\left[ a {-} \mathbf{m}(s), \mathbf{0}\right]$ and $ \mathbf{a}_{+} {=} \max\left[ a{-} \mathbf{m}(s), \mathbf{0}\right]$ (both are element-wise operations). The expectation of $f^\nnw_\text{G}$ under the policy can be easily derived from the properties of products of Gaussian densities for one component $i$ of the action vector:
\begin{eqnarray}
\mathbb{E}_{a'\sim\pi}\left[ e^{ -\frac{1}{2} u_{+,i}^\intercal ~L_{+,i}\inv(s)~u_{+,i}  -\frac{1}{2}u_{-,i}^\intercal ~L_{-,i}\inv(s) ~u_{-,i} }\right] = \hspace{0.4cm} \nonumber \\ \frac{\sqrt{\frac{L_{+,i}(s)}{L_{+,i}(s)+\Sigma_i(s)}} + \sqrt{\frac{L_{-,i}(s)}{L_{-,i}(s)+\Sigma_i(s)}}}{2}
\end{eqnarray}
Here $|\cdot|$ denotes a determinant and we note that we exploited the symmetry of the Gaussian policy around the mean. Because $\boldsymbol{\Sigma}$, $\mathbf{L}_{+}$, and $\mathbf{L}_{-}$ are all diagonal, we obtain:
\begin{eqnarray}
\mathbb{E}_{a'\sim\pi}\left[f^\nnw_\text{G}(s, a')\right] = \hspace{4.7cm} \nonumber \\ K(s) \prod_{i=1}^{d_A} \frac{\sqrt{\frac{L_{+,i}(s)}{L_{+,i}(s)+\Sigma_i(s)}} {+} \sqrt{\frac{L_{-,i}(s)}{L_{-,i}(s)+\Sigma_i(s)}}}{2}
\end{eqnarray}
We note that all these parameterizations are differentiable.

The first parameterization $f^\nnw_\text{Q}$ requires $(d_A^2 + d_A)/2$ additional network outputs, corresponding to the entries of the lower triangular matrix $\mathbf{L}_\text{Q}$. The second parameterization requires one MLP output for $K(s)$ and $d_A$ outputs for each diagonal matrix $\mathbf{L}_{+}$ and  $\mathbf{L}_{-}$. For example, for the second parameterization, given a state $s$, a single MLP computes in total  $\mathbf{m}$, $\boldsymbol{\Sigma}$, $V$, $K$, $\mathbf{L}_{+}$ and $\mathbf{L}_{-}$. The quadratic complexity of $f^\nnw_\text{Q}$ affects the computational cost of learning tasks with high-dimensional action spaces (e.g. it requires 153 parameters for the 17-dimensional Humanoid tasks of OpenAI Gym, against the 35 of $f^\nnw_\text{G}$). Finally, in order to preserve bijection between $\mathbf{L}_\text{Q}$ and $\mathbf{L}_\text{Q} \mathbf{L}_\text{Q}^\intercal$, the diagonal terms are mapped to $\mathbb{R}^+$ with a Softplus rectifier. Similarly, to ensure concavity of $f^\nnw_\text{G}$,  the network outputs corresponsing to $K$, $\mathbf{L}_{+}$ and $\mathbf{L}_{-}$ are mapped onto $\mathbb{R}^+$ by a Softplus rectifier.

The parameterization coefficients are updated to minimize the L2 error from $ \hat{Q}^\text{ret}_t$:
\begin{equation}
\mathcal{L}^\text{adv}(\nnw) =\frac{1}{2}
\mathrel{\raisebox{3pt}{$ \mathop{\mathbb{E}}_{\substack{s_k\sim B(s) \\ a_k \sim \mu_k}} $}} 
\left[ \rho_k \left( A^\nnw(s_k, a_k) {+}V^\nnw(s_k) -\hat{Q}^\text{ret}_k \right) \right] ^2 \label{eq:Atgt}
\end{equation}
Here, $\rho_t$ reduces the weight of estimation errors for unlikely actions, where $A^\nnw$ is expected to be less accurate.

Beside increasing the number of network outputs, the introduction of a parameterized $A^\nnw$ affects how the value estimators are computed (i.e. we do not approximate $Q^\nnw = V^\nnw$ when updating Retrace as discussed in Sec.~\ref{sec:pre}). This change may decrease the variance of the value estimators, but its observed benefits are negligible when compared to other HP changes. The minor performance improvements allowed by the introduction of a closed-form $A^\nnw$ parameterization are outweighed in most cases by the increased simplicity of the original {\tt V-RACER} architecture.

\fi
\end{document}